%% file: main.tex
\documentclass[11pt]{article}
\usepackage[preprint]{acl}
\usepackage{times}
\usepackage{latexsym}
\usepackage[T1]{fontenc}
\usepackage[utf8]{inputenc}
\usepackage{microtype}
\usepackage{graphicx}
\input{acl_preamble}

\title{AgentOPSD: Turn-Level Credit Assignment via Recursive Belief Revision}

\author{%
  \textbf{Yizhuo Li}\thanks{This manuscript is a rewritten, single-environment
  condensation of our earlier technical report (arXiv:2608.05987v1). The ALFWorld
  results reported here are reproduced from that work.}\\[3pt]
  Beijing Institute of Technology
}

\begin{document}
\maketitle
\input{paper_body}
\end{document}

%% file: acl_preamble.tex
\usepackage{booktabs}
\usepackage{pifont}

\newcommand{\methodname}{\text{AgentOPSD}\xspace}
\usepackage{xspace}
\usepackage{amsthm}
\newtheorem{proposition}{Proposition}
\usepackage{algorithm}
\usepackage{algpseudocode}
\makeatletter
\@ifundefined{algorithm*}{%
  \newenvironment{algorithm*}{\@dblfloat{algorithm}}{\end@dblfloat}%
}{}
\makeatother
\usepackage{url}
\usepackage{amsmath}
\usepackage{amssymb}
\usepackage{array}
\usepackage{xcolor}
\usepackage{colortbl}
\usepackage{soul}       
\definecolor{topcolor}{RGB}{252, 236, 196}
\definecolor{secondcolor}{RGB}{223, 235, 253}
\usepackage[most,skins,theorems]{tcolorbox}
\definecolor{darkgreen}{RGB}{0,128,0}

\newtcolorbox{templatebox}[1]{
  enhanced,
  unbreakable,
  colback=white,
  colframe=black!65,
  colbacktitle=black!80,
  coltitle=white,
  boxrule=0.9pt,
  arc=2pt,
  left=6pt,
  right=6pt,
  top=6pt,
  bottom=6pt,
  title={#1},
  fonttitle=\bfseries,
  sharp corners,
  boxed title style={sharp corners, boxrule=0pt}
}
\usepackage{wrapfig}

%% file: paper_body.tex
\begin{abstract}
Reinforcement learning with verifiable rewards ordinarily spreads a single
advantage over a whole trajectory, a granularity too coarse to isolate the
few decisions on which success in a long, multi-turn interaction actually
turns. Privileged self-distillation supplies denser supervision, but it
leaves unanswered how a set of local teacher--student discrepancies should be
converted into credit along a sequence. We introduce \textbf{\methodname{}},
a critic-free scheme that assigns credit recursively, one interaction turn at
a time. Inside a turn, the token-level log-probability gaps between a
skill-conditioned self-teacher and the student are summed, and that sum acts
as evidence in a recursive Bayesian update carried out in log-odds space; the
resulting revisions reallocate the outcome-derived advantage across turns
through a multiplier that is bounded and preserves its sign. With Qwen2.5-3B
and Qwen2.5-7B as policy backbones on ALFWorld, the method improves
consistently on GRPO and on self-distillation baselines, and attains
$89.1\%$ success at the 7B scale. Ablations attribute the gain to aggregating
evidence at turn boundaries and to making the belief revision depend on
history, rather than to the local gap by itself.
\end{abstract}

\begin{figure*}[t]
\centering
\includegraphics[width=\textwidth]{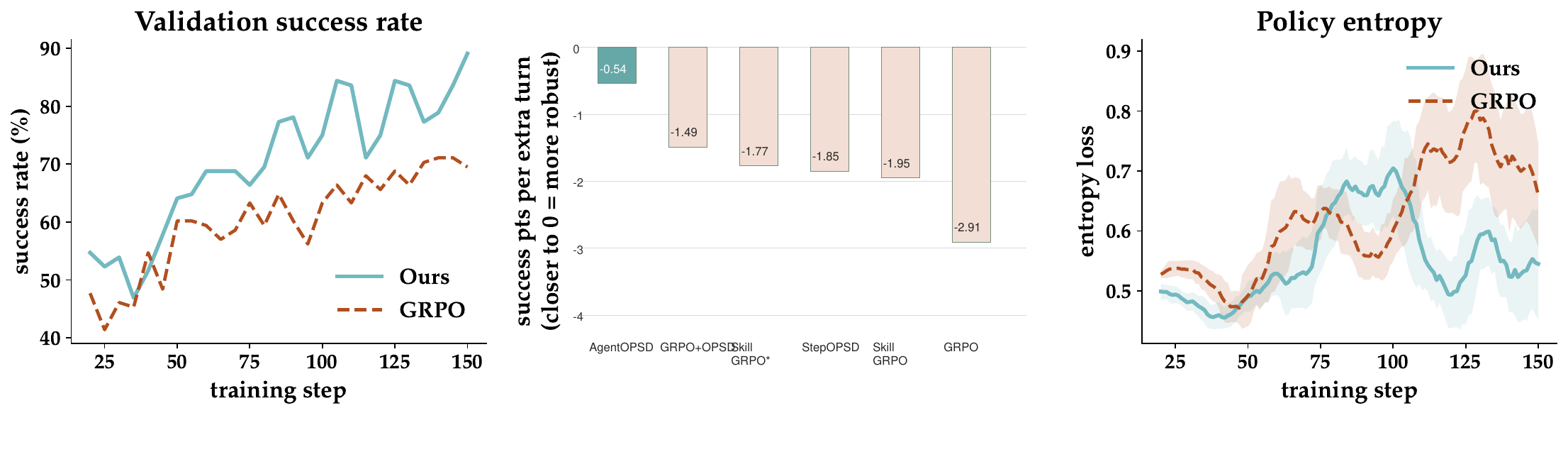}
\caption{\textbf{Training dynamics and robustness to horizon of \methodname{}}
on Qwen2.5-7B-Instruct / ALFWorld.
\textbf{(a)} Validation success rate over the course of training.
\textbf{(b)} How accuracy decays as the task horizon grows, quantified by OLS
slopes of per-sub-task success against the mean number of turns taken by
successful episodes.
\textbf{(c)} Policy entropy over the course of training.}
\label{fig:dynamics_compare}
\end{figure*}

\section{Introduction}

\label{sec:introduction}

Teaching large language models to act as agents has become a principal avenue for extending their capacity on difficult tasks~\citep{guo2025ds-r1,team2025kimi,yang2025qwen3,comanici2025gemini,team2026longcat-2601}. The setting is not static single-turn reasoning: an agent inhabits a partially observable world and keeps acting, choosing from what it currently sees and then observing the world's response before deciding again~\citep{shen2023hugginggpt,shi2025toollearning,jimenez2023swebench}. Verification, moreover, often becomes possible only once a trajectory has finished, so a training procedure must infer the role of one intermediate step from a single sparse terminal signal. Yet steps inside a trajectory are far from interchangeable: a successful run can contain redundant, spurious or misleading actions, and a failing run can contain sensible reasoning.

Group-relative policy optimization, with GRPO~\citep{shao2024deepseekmath,yu2025dapo} being the canonical instance alongside its agentic descendants~\citep{dong2025arpo,feng2025gigpo}, derives a single advantage for the whole trajectory from the outcome reward and applies it unchanged at every position. Credit that is spread this flatly cannot separate a handful of pivotal decisions from routine operations around them, and the distortion worsens as turns accumulate. Per-turn credit is therefore a prerequisite both for locating decisions that actually drive the outcome and for supervising long horizons at finer granularity.

Denser token-level supervision has been pursued along a separate line. In on-policy distillation~\citep{ye2026opcd,yang2026g-opd,coreteam2026mimov2}, a student is trained under a teacher on rollouts of its own, while the \emph{self}-distillation variants of this idea~\citep{zhao2026opsd,he2026sdzero} do away with any separate teacher by conditioning one and the same policy on privileged information that exists only during training~\citep{lu2026skill0}. Reinforcement learning has likewise used such signals as an auxiliary supervision source.

Two mismatches appear once OPSD meets agentic reinforcement learning. The first concerns granularity: token-level signals sit awkwardly against an interaction in which many tokens jointly constitute one action and the environment reacts only when a turn ends. The second concerns memory: even step-aware methods score each turn separately~\citep{zhang2026stepopsd}, leaving earlier evidence unused. What must be solved, then, is how local OPSD signals become turn-level credit that depends on history.

We begin from the position that \textbf{what should govern a turn's credit is not its local signal taken on its own, but how far that signal moves the estimated probability of eventual success}. Precision comes from reading the per-turn self-distillation gap as newly arrived evidence driving a Bayesian belief update~\citep{astrom1965optimal,kaelbling1998planning}. The belief state is then the probability of eventual task success given the history so far.

This observation leads us to \methodname{} (\emph{agentic on-policy self-distillation}), a method that assigns credit turn by turn in long-horizon agents. Token-level teacher--student log-probability gaps are first pooled by \methodname{} into turn-level evidence. A Bayesian belief state, initialized at the mean group success rate, is then revised recursively turn after turn in log-odds space, with neither extra rollouts nor a learned critic required. The verifier fixes the direction of optimization globally, while the belief updates redistribute the trajectory-level learning signal over individual turns. \methodname{} is evaluated on ALFWorld~\citep{shridhar2020alfworld} at two model scales, and Figure~\ref{fig:dynamics_compare} shows it to outperform GRPO and strong self-distillation baselines consistently. The ablations point to two sources of the improvement: pooling token-level signals at the turn boundaries that coincide with environment transitions, and reframing a gap that stands on its own as a revision of belief that recurs along the trajectory.

We summarize the contributions of this work below:

\begin{itemize}

\item Credit at the level of a turn is cast as the amount by which one turn revises the estimated chance that the run ends in success. Carrying the recursion in log-odds space ties the evidence of one turn to a Bayesian update applied repeatedly, and makes plain that a self-distillation gap on its own does not amount to sequential credit.

\item We propose \methodname{}, which collapses teacher--student log-probability gaps at the token level into turn-level evidence that lines up with environment transitions, then carries it recursively through the belief over trajectory success; neither extra rollouts nor a learned critic is required.

\item On ALFWorld, at two model scales, experiments together with ablations show \methodname{} to outperform GRPO and strong self-distillation baselines consistently, and the ablations confirm that aggregating at turn boundaries and revising belief by recursion bring benefits that are complementary.

\end{itemize}

\begin{figure*}[t]
\centering
\includegraphics[width=\textwidth]{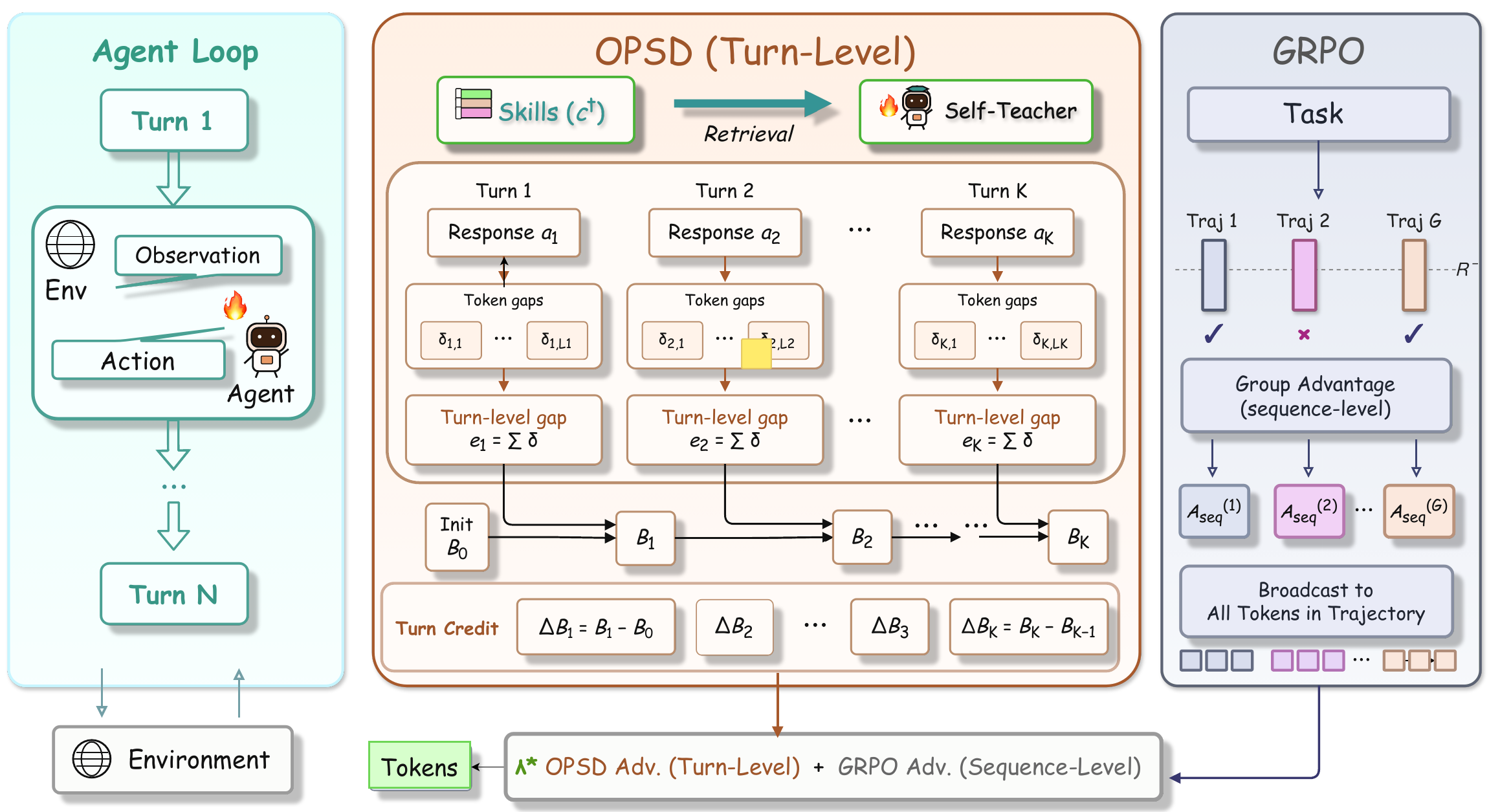}
\caption{\textbf{An overview of \methodname{}.} \textbf{Left:} the loop in which the agent acts, exchanging with the environment across turns $1,\dots,K$. \textbf{Middle:} in three steps, \methodname{} turns the single sequence-level advantage of GRPO into advantages reshaped at the turn level: \textbf{(1)} the token-level teacher--student gaps $\delta_{k,t}$ are aggregated inside a turn, giving the turn-level gap $e_k$; \textbf{(2)} a belief state $B_k$, started from the group success rate, is updated recursively, and its marginal revision $\Delta B_k = B_k - B_{k-1}$ is read off; \textbf{(3)} the sequence-level advantage $A^{(i)}_{seq}$ is reshaped for each turn as $\tilde{A}^{(i)}_k$. \textbf{Right:} vanilla GRPO instead sends one and the same $A^{(i)}_{seq}$ to all tokens and turns. Every token of turn $k$ therefore takes on $\tilde{A}_k$.}
\label{fig:overview}
\end{figure*}

\section{Methodology}
\label{sec:method}

\subsection{Task, Interaction, and Notation}
\label{sec:setup}

Each episode opens with a task $x$ and an initial observation $o_0$, so that the interaction starts from the state $s_1=(x,o_0)$. At turn $k$ the policy $\pi_\theta$ emits an action conditioned on the interaction history $s_k$ that the agent can currently see; we denote that action by $a_k$, its $t$-th token by $y_{k,t}$, and the token count of the action by $L_k$. The observation $o_k$ returned by the environment then extends the history through $s_{k+1}=(s_k,a_k,o_k)$. Chaining these steps over $K$ turns produces a trajectory $\boldsymbol{\tau}=(s_1,a_1,o_1,\ldots,s_K,a_K,o_K)$, to which a verifier attaches a binary outcome reward $R(\boldsymbol{\tau})$.
\begin{equation}
a_k=(y_{k,1},\ldots,y_{k,L_k})\sim\pi_\theta(\cdot\mid s_k),
\end{equation}
Every task is handled by drawing $G$ trajectories under group-relative policy optimization, from whose rewards a sequence-level advantage is derived. Throughout, the index $i$ runs over these $G$ samples, $\bar R$ and $\widehat{\sigma}_R$ are the mean reward of the group and its standard deviation, and the small positive constant $\epsilon_0$ recurs everywhere as a guard against division by zero and against unbounded log-odds. The resulting advantage $A_{\mathrm{seq}}^{(i)}$ is handed to all tokens of trajectory $i$ in exactly the same form, so no distinction of credit between turns is ever drawn.
\begin{equation}
A_{\mathrm{seq}}^{(i)}
=
\frac{R^{(i)}-\bar R}{\widehat{\sigma}_R+\epsilon_0},
\qquad
\bar R=\frac{1}{G}\sum_{j=1}^{G}R^{(j)}.
\label{eq:grpo_adv}
\end{equation}

\subsection{Turning Outcome Influence into Bayesian Turn Evidence}
\label{sec:bayesian_evidence}

The counterfactual influence of turn $k$ could be quantified head-on only by averaging the outcome reward over all continuations that might follow $a_k$, and for interactions spanning many turns that marginalization is hopeless. Our stance is instead retrospective and evidential. Let $C$ denote the event in which the trajectory ends successfully. An action that advances the task should occur more often inside successful behavior than inside unsuccessful behavior, and Bayes' rule turns this observation into a change of belief about $C$, captured by a likelihood ratio on the action side~\citep{astrom1965optimal,kaelbling1998planning}:
\begin{equation}
\begin{aligned}
&\operatorname{logit}p(C\mid s_k,a_k)-\operatorname{logit}p(C\mid s_k)\\
&\qquad=\log\frac{p(a_k\mid s_k,C)}{p(a_k\mid s_k,\neg C)}.
\end{aligned}
\label{eq:bayes_turn}
\end{equation}
The map $\operatorname{logit}(u)=\log\frac{u}{1-u}$ converts a probability into log-odds. What sits on the right is the ideal Bayes factor~\citep{kass1995bayesfactors}, i.e., the ratio between the likelihood of $a_k$ under success and its likelihood under failure; its sign tells us at once whether the turn adds to the case for eventual success or subtracts from it.

Neither outcome-conditional behavioral distribution can be inspected directly, so the Bayesian evidence has to be recovered after the fact. We obtain it from a per-turn self-distillation contrast that is actually computable: a privileged self-teacher, associated with success, is set against the ordinary student policy. The two share the parameters $\theta$ and both score the same action, as sampled from the student~\citep{zhao2026opsd}. Such a contrast supplies a tractable stand-in for a belief update that would otherwise stay out of reach, and the quality of the stand-in, together with the assumptions that justify it, is examined below. The two branches are told apart only by the token context they receive:
\begin{equation}
h_{k,t}=(s_k,y_{k,<t}),
\qquad
h_{k,t}^{+}=(s_k,c^+,y_{k,<t}).
\end{equation}
In this notation $y_{k,<t}$ is the prefix of tokens already emitted within turn $k$, while $c^+$ is a skill retrieved at training time only, summarizing subgoals and action patterns of use for the task~\citep{xia2026skillrl}. The skill-conditioned branch stands in for success-associated behavior, whereas the branch that sees no skill supplies the background likelihood.

For a generated token $y_{k,t}$, the detached likelihood contrast is written as
\begin{equation}
\delta_{k,t}
=
\log\pi_\theta(y_{k,t}\mid h_{k,t}^{+})
-
\log\pi_\theta(y_{k,t}\mid h_{k,t}).
\label{eq:gap}
\end{equation}
Whenever $\delta_{k,t}$ is positive, the skill has made the generated token more probable than it would otherwise be. Summing the contrast over the $L_k$ tokens that make up the turn gives the turn-level evidence
\begin{equation}
e_k
=
\sum_{t=1}^{L_k}\delta_{k,t}
=
\log
\frac{\pi_\theta(a_k\mid s_k,c^+)}
     {\pi_\theta(a_k\mid s_k)}.
\label{eq:turn_evidence}
\end{equation}
The quantity $e_k$ therefore acts as a computable, retrospective surrogate for the ideal turn evidence of Eq.~\eqref{eq:bayes_turn}; Appendix~\ref{app:bayes_approx} states the conditions under which the two agree. Bayes' rule also permits a more general identity:
\begin{equation}
\log\frac{p(a_k\mid s_k,C)}
          {p(a_k\mid s_k)}
=
\log\frac{p(C\mid s_k,a_k)}
          {p(C\mid s_k)}.
\label{eq:bayes_evidence_proxy}
\end{equation}
Read this way, $e_k$ is an evidential score, and its sign records the direction in which $a_k$ moves the support for eventual success. Sign agreement of this kind is the very property that \methodname{} draws on, which is why $e_k$ is treated here as an evidence proxy that is both tractable and Bayesian in spirit.

\subsection{Recursive Evidence State}
\label{sec:recursive_support}

Taken on its own, a local score $e_k$ cannot tell whether the evidence it carries is decisive or whether earlier turns have already made it redundant. We therefore keep an accumulator of evidence that decays over turns, and score each turn according to how far it moves the support state maintained so far:
\begin{equation}
\begin{aligned}
B_0&=\operatorname{clip}(\bar R,\epsilon_0,1-\epsilon_0),
\qquad c_0=0,\\
c_k&=\gamma\,c_{k-1}+e_k,\\
\ell_k&=\operatorname{logit}(B_0)+c_k\\
&=\operatorname{logit}(B_0)+\sum_{j=1}^{k}\gamma^{\,k-j}e_j,
\end{aligned}
\label{eq:belief}
\end{equation}
with $B_k=\sigma(\ell_k)$ and $\sigma(u)=(1+e^{-u})^{-1}$. The seed value is the proportion of successful trajectories in a group of size $G$, namely $\bar R=S/G$, which is exactly the group mean GRPO already forms (Prop.~\ref{prop:v0}); $B_0$ is then obtained by clipping that fraction into $[\epsilon_0,1-\epsilon_0]$ with $\epsilon_0{=}10^{-4}$, and the resulting log-odds remain finite even for a group that is uniformly correct or uniformly wrong. Evidence is gathered in $c_k$, and $\gamma\in(0,1]$ discounts it geometrically as turns recede; the discount touches $c_k$ alone, since the prior $\operatorname{logit}(B_0)$ enters every step unchanged. Setting $\gamma{=}1$ gives back the plain, undiscounted sum of log-likelihood ratios familiar from sequential testing~\citep{wald1945sequential}, whereas $\gamma{<}1$ weights recent turns more heavily, so evidence gathered many turns ago can no longer pin the support level. Because $e_k$ is produced by the self-teacher (\S\ref{sec:bayesian_evidence}), $B_k$ is read as a relative level of support rather than as a calibrated probability of success.

Turn importance is then defined through the marginal revision that a turn induces in this support:
\begin{equation}
\begin{aligned}
\Delta B_k
&=B_k-B_{k-1}
=\sigma(\ell_k)-\sigma(\ell_{k-1}),\\
\Delta B_k
&\approx B_{k-1}(1-B_{k-1})\big(e_k-(1-\gamma)\,c_{k-1}\big) .
\end{aligned}
\label{eq:belief_revision}
\end{equation}
What the increment $\ell_k-\ell_{k-1}=e_k-(1-\gamma)c_{k-1}$ contains is the newly arrived evidence once the decayed carry-over has been removed, and the increment is multiplied by the state sensitivity $B_{k-1}(1-B_{k-1})$: evidence counts for most while support is still uncertain, and its effect fades as support approaches saturation. At $\gamma{=}1$ the expression reduces to $B_{k-1}(1-B_{k-1})\,e_k$. We apply the recursion at turn boundaries, keeping a token-level variant purely as an ablation.

\paragraph{Credit oriented by the outcome.}
\label{sec:outcome_credit}

The revision becomes outcome-aware once it is oriented by the terminal update, after which both its size and its direction can be read off:
\begin{equation}
q_k
=
\operatorname{sign}(A_{\mathrm{seq}})\,\Delta B_k.
\label{eq:credit}
\end{equation}
The \emph{magnitude} $|\Delta B_k|=|q_k|$ measures how large a revision the turn makes, while the \emph{sign} $\operatorname{sign}(q_k)$ records whether that revision matches the outcome signal supplied by the verifier. After the within-trajectory standardization described below, turns scoring above the average $q_k$ are amplified and turns scoring below it are attenuated; the multiplier itself stays strictly positive, so the direction of the GRPO update is never reversed.

\subsection{Bounded Multiplicative Reshaping}
\label{sec:reshape}

What the raw credit $q_k$ is allowed to touch is the magnitude of the advantage derived from the verifier, and nothing else. Trajectory $i$ contributes $K_i$ turn-wise credits; we standardize them and pass the result through a bounded multiplier:
\begin{equation}
\begin{aligned}
\mu_q^{(i)}&=K_i^{-1}\sum_{j=1}^{K_i}q_j^{(i)},\\
\sigma_q^{(i)}&=\sqrt{K_i^{-1}\sum_{j=1}^{K_i}\big(q_j^{(i)}-\mu_q^{(i)}\big)^2},\\
z_k^{(i)}&=\frac{q_k^{(i)}-\mu_q^{(i)}}{\sigma_q^{(i)}+\epsilon_0},\\
w_k^{(i)}&=\operatorname{clip}\!\left(1+bz_k^{(i)},\,1-b,\,1+b\right),\\
\widetilde A_k^{(i)}&=A_{\mathrm{seq}}^{(i)}\big[(1-\lambda)+\lambda w_k^{(i)}\big],\\
&\qquad b\in(0,1),\quad\lambda\in[0,1].
\end{aligned}
\label{eq:bounded_reshape}
\end{equation}
Here $\mu_q^{(i)}$ is the average credit inside trajectory $i$ and $\sigma_q^{(i)}$ its standard deviation, $z_k^{(i)}$ is the credit after standardization, with $\epsilon_0$ keeping the normalization stable, the constant $b$ confines $w_k^{(i)}$ to $[1-b,1+b]$, and $\lambda$ sets how much reshaping is applied.

Every token $t$ inherits the advantage of its own turn, $\widetilde A_{\kappa_i(t)}^{(i)}$, so that the training objective reads
\begin{equation}
\begin{aligned}
\mathcal{L}_{\methodname}(\theta)
&=-\frac{1}{G}\sum_{i=1}^{G}\frac{1}{\sum_tM_{i,t}}\\
&\quad\sum_tM_{i,t}\,\Phi_{i,t}\\
&\quad+\beta\mathcal{L}_{\mathrm{KL}},\\
\Phi_{i,t}&=\min\!\Big(r_{i,t}\widetilde A_{\kappa_i(t)}^{(i)},\\
&\quad
\operatorname{clip}(r_{i,t},1-\varepsilon,1+\varepsilon)\,
\widetilde A_{\kappa_i(t)}^{(i)}\Big),\\
r_{i,t}&=\frac{\pi_\theta(y_{i,t}\mid h_{i,t})}{\pi_{\theta_{\mathrm{old}}}(y_{i,t}\mid h_{i,t})}.
\end{aligned}
\label{eq:agentopsd_objective}
\end{equation}
In this expression $\Phi_{i,t}$ is the clipped per-token surrogate, the mask $M_{i,t}\in\{0,1\}$ selects the response tokens that count, $\kappa_i(t)$ returns the turn to which token $t$ belongs, $r_{i,t}$ is the importance ratio between the current policy and the rollout policy $\pi_{\theta_{\mathrm{old}}}$, $\varepsilon$ fixes the clipping radius, and $\beta$ scales the final term. Nothing is added in the way of a separate distillation loss: the detached signal from the self-teacher reaches the objective through $\widetilde A$ and through nothing else.

\input{tables/experiment}

\section{Experiments}

\subsection{Experimental Setup}
\label{sec:experiments}

\paragraph{Benchmark.} Every experiment reported here is run on
\textit{ALFWorld}~\citep{shridhar2020alfworld}, a benchmark in which embodied agents act through
text. Its task suite is organized into six families of household activity: Pick and Place (Pick),
Pick Two and Place (Pick2), Pick Clean then Place (Clean), Pick Cool then Place (Cool), Pick Heat
then Place (Heat), and Look at Object in Light (Look).

\paragraph{Training configuration.} Qwen2.5-3B/7B-Instruct models are trained with
$8\times$H800 GPUs. Privileged skills are drawn from the \texttt{SkillBank} released with
SkillRL~\citep{xia2026skillrl} and matched by keywords; they are supplied during training alone,
and no external skill is consulted at inference. The prior $B_0$ is initialized to the share of
successful rollouts within each GRPO group, which coincides with the usual group mean $\bar R$.
Every remaining optimization detail is inherited from our common protocol. Full tables of training
and \methodname{}-specific hyperparameters are given in
Table~\ref{tab:hyperparams} of Appendix~\ref{app:hyperparams}.

\paragraph{Compared methods.} Three families of baselines enter the comparison.
\textbf{(1) No training:} \textit{Vanilla}, i.e.\ the backbone as released, and \textit{Skill-Prompt},
which concatenates retrieved skills onto the inference-time context.
\textbf{(2) Group-relative RL:} \textit{GRPO}~\citep{shao2024deepseekmath}; \textit{Skill-GRPO},
for which skills are added to the training prompt and evaluation is run either with them
(\textit{Skill-GRPO*}) or without.
\textbf{(3) Self-distillation RL:} \textit{OPSD}~\citep{zhao2026opsd},
\textit{StepOPSD}~\citep{zhang2026stepopsd}, and their combination \textit{GRPO+OPSD}; these three
differ in the route by which the teacher--student gap reaches the policy. Backbone, data, and
training budget are held identical across every row, and complete algorithmic descriptions are
deferred to Appendix~\ref{appendix:algorithm}.

\subsection{Main Results}

\paragraph{Gains trace to credit construction, not to privileged information.}
\methodname{} and the privileged baselines consume an identical pool of retrieved skills, so what
separates them is chiefly the way a skill-induced teacher--student discrepancy is converted into a
learning signal. On ALFWorld, \methodname{} beats every baseline it is compared against at both
model scales. Holding the available information constant in this fashion localizes the origin of
the improvement: a gap measured locally between teacher and student does not by itself amount to a
dependable credit signal. Once that gap is accumulated into a belief state and credit is granted
according to how the belief is revised, the turns that actually alter the predicted outcome are
pinned down far more precisely.

\paragraph{The margin widens as the interaction horizon lengthens.}
Because \methodname{} is aimed at the setting in which a uniformly assigned credit does the
greatest harm, a second question follows: how quickly does accuracy fall away when tasks need more
turns? In Figure~\ref{fig:dynamics_compare}(b), success on each sub-task is regressed against the
empirical mean turn count of successful episodes on ALFWorld with Qwen2.5-7B, so that a fitted slope
reads as the number of success points surrendered for each extra turn. GRPO, which broadcasts one
advantage to all turns alike, sheds $-2.91$ points per turn, whereas \methodname{} stays markedly
flatter at $-0.54$. Such a gap is what turn-level credit predicts: every added decision in a long trajectory
must otherwise share a single broadcast advantage, so a revision conditioned on history pays off
more as the horizon stretches.

\subsection{Ablating the Mechanism Components}
\label{sec:mechanism_ablation}

Table~\ref{tab:ablation} takes up each design decision separately on ALFWorld with Qwen2.5-7B,
removing or substituting one ingredient per row. With every ingredient present, success reaches
$89.1$.
\input{tables/ablation}
\paragraph{Turn-level granularity and recursive revision.}
Belief tracking at the level of turns preserves the full-method score; when accumulation is instead
done token by token, success drops to $85.9$. Feedback from the environment is delivered for a whole
action and never for an isolated token, so splitting a decision into token-sized units loosens the
tie between the gap and the outcome. A further decline, to $82.8$, follows from discarding the
recursive revision $\Delta B_k$ in favour of the raw local gap $e_k$. The raw quantity scores every
turn on its own, whereas $\Delta B_k$ records the change that the gap induces in the belief
assembled from all preceding turns; a local gap that looks decisive while the outcome is still
undecided therefore turns redundant once the accumulated state already indicates how the episode
will end. Isolating the recursion in this controlled manner substantiates the central claim of the
paper, namely that a local gap is not yet sequential credit.
\paragraph{Sign alignment with the verifier outcome.}
If the sign is dropped and only the magnitude $|\Delta B_k|$ is kept (Eq.~\ref{eq:credit}), that is,
if $|\Delta B_k|$ is standardized in place of the signed $q_k$, success falls back to $80.5$. A
magnitude reveals that the belief state has shifted, yet it cannot say whether the shift agrees with
the outcome returned by the verifier: one and the same upward revision is consistent with success
and inconsistent with failure. Making the direction explicit separates those two cases, so that
revisions agreeing with the outcome are rewarded more and revisions contradicting it are rewarded
less.

\paragraph{Anchoring to the state prior.}
Discarding the empirical prior term
$B_0=\operatorname{clip}(\bar R,\epsilon_0,1-\epsilon_0)$ brings success down to $78.9$. Before
any trajectory-specific gap accumulates, the group success rate $\bar R$ already supplies an estimate
of task difficulty grounded in the verifier; beyond that, $B_0$ sets the initial log-odds and
therefore the region in which the $B(1-B)$ gate operates. Lacking such an anchor, the starting
uncertainty of a trajectory is arbitrary, early revisions of the belief may be rescaled wrongly, and
the early turns that look pivotal can be misjudged. Read together, the three ablations assign
distinct roles: localizing credit is the task of belief revision, agreeing with the final outcome is
the task of the signed direction, and holding the reference point steady is the task of prior
anchoring.

\subsection{Hyperparameter Sensitivity}
\label{sec:sensitivity}

\begin{figure*}[t]
\centering
\includegraphics[width=\textwidth]{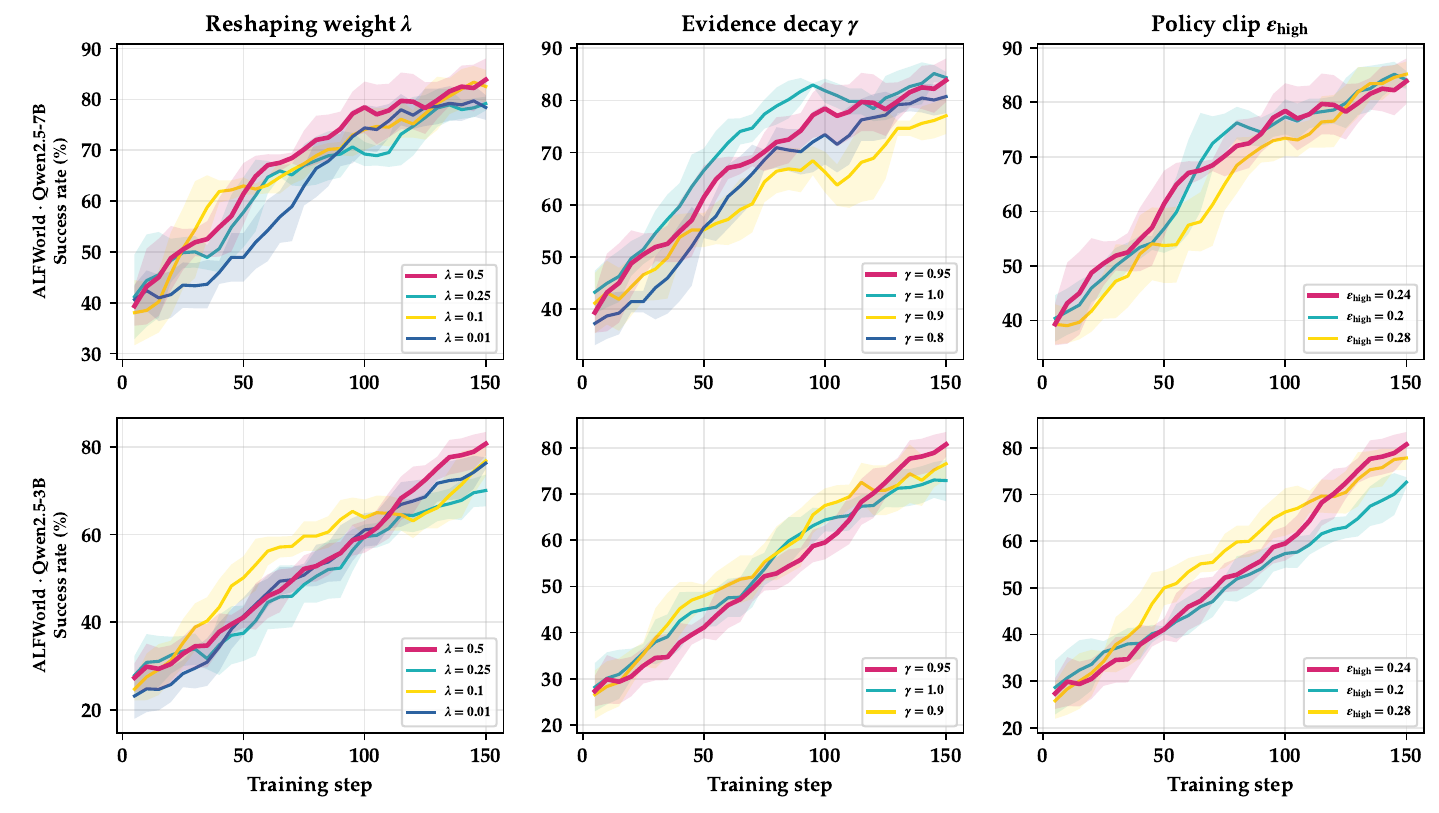}
\caption{\textbf{Sensitivity of \methodname{} to its hyperparameters.} Rows correspond to ALFWorld
with Qwen2.5-7B and ALFWorld with Qwen2.5-3B. Each column varies a single knob ($\lambda$, $\gamma$,
$\epsilon_{\mathrm{high}}$) while the other two keep our values.
Each curve is a rolling mean, and the shaded bands give a local $\pm1$ standard deviation.}
\label{fig:sensitivity}
\end{figure*}

The mechanism ablation has established that each component is necessary; the question here is how
far \methodname{} depends on the continuous hyperparameters. One knob is varied while all others stay
at the full-\methodname{} values
($\lambda{=}0.5$, $\gamma{=}0.95$, $\epsilon_{\mathrm{high}}{=}0.24$; Appendix~\ref{app:hyperparams}),
and two configurations are examined in Figure~\ref{fig:sensitivity}, namely Qwen2.5-7B on ALFWorld
($89.1$) and Qwen2.5-3B on ALFWorld ($84.4$), with both checkpoints collected in the grid below.

\paragraph{Weight of the reshaping term $\lambda$.}
The sweep $\lambda\in\{0.5,0.25,0.1,0.01\}$ moves the objective from pure GRPO
($\lambda{=}0$) towards fully reshaped beliefs. Of all the knobs this one matters most: $\lambda{=}0.5$
gives the best result, and every smaller value costs accuracy ($89.1$ at $\lambda{=}0.5$ versus
$84.4/85.9/83.6$), as expected when a smaller $\lambda$ weakens the bounded multiplier until
turn-level credit is effectively discarded. We therefore keep $\lambda{=}0.5$ everywhere.
\paragraph{Decay of accumulated evidence $\gamma$.}
Turn-level evidence is accumulated geometrically, $c_k=\gamma\,c_{k-1}+e_k$; in log-odds form,
$\ell_k$ is the offset $\operatorname{logit}(B_0)$ plus the discounted sum
$\sum_{j\le k}\gamma^{\,k-j}e_j$.
For $\gamma\in\{1.0,0.95,0.9,0.8\}$ the result moves by only a handful of points
($87.5/82.0/85.2$) with no monotone ordering, which indicates that the recursion does not hinge on
the rate at which older evidence is discounted. All main results consequently adopt the mild choice
$\gamma{=}0.95$.
\paragraph{Clipping of the policy ratio $\epsilon_{\mathrm{high}}$.}
Holding $\epsilon_{\mathrm{low}}{=}0.2$ fixed while
$\epsilon_{\mathrm{high}}\in\{0.2,0.24,0.28\}$ is varied (clip-higher~\citep{yu2025dapo}),
\methodname{} barely moves ($88.3$ for both $0.2$ and $0.28$), suggesting that the reshaped objective
retains the trust-region robustness of GRPO. Taken as a whole, $\lambda$ is the sole knob whose
effect is systematic; at either ALFWorld scale examined here the remaining knobs stay within a few
points of the full configuration, so \methodname{} is insensitive to these secondary
hyperparameters.

\section{Related Work}

\subsection{Reinforcement Learning for Long-Horizon Agents}

Work on RL with verifiable rewards initially targeted single-turn
reasoning~\citep{shao2024deepseekmath,guo2025ds-r1,yu2025dapo,lu2026uir1},
then progressed toward agents that act across many steps in embodied
text environments~\citep{shridhar2020alfworld,lu2025uis1}. Only a sparse
terminal reward is available in such settings, so identifying the turns
that merit credit becomes particularly hard. Vanilla GRPO avoids this
question altogether: it spreads one trajectory-level advantage evenly
across all turns. GiGPO~\citep{feng2025gigpo} improves matters on the
reward side by pairing episode-level advantages with step-level
counterparts estimated from anchor states that reappear across
trajectories. \methodname{}, by contrast, grounds its turn-level
evidence in privileged teacher--student likelihood gaps, and
apportions credit through recursive updates of a belief state. The two
approaches therefore rely on complementary sources of
information---environment rewards in the case of GiGPO,
self-distillation evidence in the case of \methodname{}.

\subsection{Distilling On-Policy Signals}

In on-policy distillation, a policy is optimized on its own rollouts
with guidance from a
teacher~\citep{agarwal2024gkd,gu2026minillm,wen2023fdivergence}. Newer
self-distillation variants drop the separate teacher entirely,
conditioning the teacher branch instead on privileged information that
is accessible only during
training~\citep{zhao2026opsd,he2026sdzero,lu2026skill0}. The
teacher--student log-probability gap produced in this way has lately
been brought into RLVR in the form of local supervision, where it may be
reweighted, annealed on a schedule, or folded into a denser
reward~\citep{xu2026tip,wang2026tcod,ye2026opcd,he2026sdzero}.
Throughout this literature the gap is mostly handled as a token-level or
step-level quantity. Yet such local signals align poorly with action
turns, whose contribution depends on evidence gathered across earlier
interactions. StepOPSD~\citep{zhang2026stepopsd} moves in this direction
by pooling the teacher--student signal across action-centered step
spans, though each span is still scored by its local log-ratio.
\methodname{} abandons this local perspective: it pools token-level gaps
within individual turns and then recursively incorporates that evidence
into a running support state.

\subsection{Assigning Credit over Long Horizons}

Assigning credit over long horizons is a problem with a long history.
PPO trains a value function and, through GAE, converts it into per-step
temporal-difference signals~\citep{schulman2017ppo,schulman2016gae}.
Under sparse, delayed rewards, two families of methods recover
intermediate structure: return-decomposition approaches such as RUDDER
push the terminal reward back onto the steps responsible for
it~\citep{arjona2019rudder}, whereas process reward models and
Monte-Carlo credit schemes like VinePPO infer intermediate value from
additional rollouts or a learned
scorer~\citep{cui2025prime,kazemnejad2024vineppo}. Both regain per-step
resolution at exactly the cost GRPO was meant to avoid---an extra
trained critic, reward model, or heavy rollout budget. \methodname{}
restores a per-turn value signal inside the critic-free, group-relative
framework, and pays only for one teacher forward pass. Here the belief
state plays a role analogous to GAE's value baseline, and its per-turn
revision parallels the TD signal, all without learning a value network.

\section{Conclusion}
This paper tackles credit assignment for long-horizon language agents,
where a trajectory-level reward can hardly separate decisive actions
from routine or redundant ones. The central claim of this work is that
a turn's credit should capture more than its local signal: it should
capture the way in which that signal revises an accumulated belief about
the trajectory's eventual success. \methodname{} realizes this idea by
aggregating token-level self-distillation gaps within turn boundaries
that follow the environment's structure, and by recursively maintaining
a belief about trajectory success in log-odds space. The resulting
belief revisions spread the trajectory-level advantage across turns,
requiring neither additional rollouts nor a learned critic. On ALFWorld and at two model scales, \methodname{}
consistently outperforms GRPO as well as strong self-distillation
baselines, and ablations verify that turn-level aggregation and
history-dependent belief revision both matter. Overall, these results
suggest that recursive belief updating offers a simple yet effective
route to recovering fine-grained temporal credit for critic-free
agentic RL.

\section{Limitations}
\label{sec:limitations}
The empirical evaluation is confined to one text-based embodied benchmark,
ALFWorld, and to two checkpoints of a single model family. We therefore make
no claim that the observed gains carry over to other environments, larger
model scales, or other modalities. The formulation presupposes that turn
boundaries can be read off the environment's interaction structure and that
a skill-conditioned self-teacher is available during training; settings with
continuous feedback or without a usable teacher signal would call for a
different construction. Group-relative advantages still require several
rollouts per prompt, so the approach does not remove the sampling cost of
GRPO-style training. The analysis in the appendix is developed under
simplifying assumptions about the belief update and does not establish
optimality or convergence of the complete training objective. Hyperparameter
sensitivity is measured through one-dimensional sweeps around a single
operating point; interactions between the knobs and behaviour at horizons
beyond those present in ALFWorld remain unexamined. Finally, success rate is
the only reported metric, and may not reflect other desiderata such as sample
efficiency or robustness to distribution shift.

\clearpage
\bibliography{references}
\clearpage

\appendix
\renewcommand{\contentsname}{Table of Contents}
\setcounter{tocdepth}{2}
\tableofcontents
\clearpage
\include{sections/proof}

\include{sections/algorithm}

\section{Datasets}
\label{app:datasets}
We run all experiments in ALFWorld, one multi-turn, text-based embodied
environment focused on household reasoning.

\paragraph{ALFWorld}~\citep{shridhar2020alfworld} contains six task
categories---Pick and Place, Look at Object in Light, Pick Clean then
Place, Pick Heat then Place, Pick Cool then Place, and Pick Two and
Place. Given a language goal together with the textual observations it
receives, the agent repeatedly chooses among the admissible actions
until the goal is achieved.

\section{Baseline Details}
\label{app:baseline_details}
We evaluate \methodname{} against baselines from three families, all
assessed under a common protocol. By default the model only observes the
standard task prompt together with the interaction history the
environment returns; a method marked with $*$ additionally receives a
retrieved skill at validation and test time. The $*$ flag therefore
marks the only departure from this shared protocol.

\paragraph{Vanilla.} Here the off-the-shelf instruction-tuned backbone
is evaluated directly, with its parameters untouched by post-training.
\paragraph{Skill-Prompt$^{*}$.} Parameters remain frozen as in Vanilla,
but a retrieved task-relevant skill is added in front of the context
during validation and testing. This isolates the inference-time benefit
of skills in the absence of any parameter update.
\paragraph{GRPO}~\citep{shao2024deepseekmath}. GRPO is a critic-free,
group-relative RL method: for every task it samples a group of
trajectories, maps their terminal rewards onto normalized relative
advantages, and then optimizes a clipped surrogate objective, assigning
the resulting sequence-level advantage uniformly to each token in the
trajectory.
\paragraph{Skill-GRPO / Skill-GRPO$^{*}$.} This variant adds a
retrieved skill to the GRPO training prompt. Skill-GRPO withholds the
skill at inference, testing whether the guidance has been internalized,
while Skill-GRPO$^{*}$ retains it.
\paragraph{OPSD}~\citep{zhao2026opsd}. An on-policy self-distillation
scheme: a teacher branch with access to privileged context during
training re-evaluates each token sampled by the student, producing dense
token-level targets through distribution matching. The teacher's outputs
are detached, while the privileged context is dropped at inference.
\paragraph{GRPO+OPSD.} Here the GRPO loss at trajectory level and the
OPSD objective at token level are optimized jointly, combining
outcome-based RL with generic self-distillation in a straightforward
way.
\paragraph{StepOPSD}~\citep{zhang2026stepopsd}. This baseline applies
the self-distillation signal at the granularity of turns (steps) instead
of tokens, and uses the local signal of each step in isolation.

All post-training baselines use the same backbone models, the ALFWorld
environment interface, the training data, and the training budget as
\methodname{}; what distinguishes them is chiefly the optimization
objective and whether retrieved skills are provided at training time or
at evaluation.

\section{Evaluation Metrics}
\label{app:metrics}
\paragraph{ALFWorld.} We measure the overall success rate on the
evaluation tasks, i.e., the proportion of episodes that attain the
specified goal; this equals a sample-weighted average across the six
per-category success rates.

\section{Hyperparameters}
\label{app:hyperparams}
Table~\ref{tab:hyperparams} lists the hyperparameters of \methodname{}
used in all of our experiments. We deliberately adopt a \emph{single}
configuration for the two ALFWorld backbone sizes (Qwen2.5-3B and
Qwen2.5-7B) rather than tuning per scale: \methodname{} works at the
granularity of turns, with reshaping weight $\lambda{=}0.5$, multiplier
band $b{=}0.2$, gap accumulation decay $\gamma{=}0.95$, policy clipping
$\epsilon_{\mathrm{low}}{=}0.2$ / $\epsilon_{\mathrm{high}}{=}0.24$, and
takes the (clipped) empirical group success rate $\bar R=S/G$ as the
prior $B_0$ of the state.
The sensitivity analysis of \S\ref{sec:sensitivity} varies $\lambda$,
$\gamma$ and $\epsilon_{\mathrm{high}}$ around this configuration and
finds that none of the swept values improves on it meaningfully, so we
keep one shared setting for both scales rather than per-scale tuning.
\begin{table*}[t]
\centering
\caption{\textbf{Hyperparameters}. Here $\eta$ is the learning rate and
$G$ the group size; $\epsilon_{\mathrm{low}}/\epsilon_{\mathrm{high}}$
give the PPO clip range, while $\alpha_{\mathrm{KL}}$ is the coefficient
of the KL penalty toward the reference policy. SRS denotes the strategy
used for skill retrieval (KM = keyword matching). The
\methodname{}-specific knobs are the reshaping weight $\lambda$
(\texttt{mult\_lambda}), the multiplier band $b$ (\texttt{mult\_band}),
and the gap-accumulation decay $\gamma$ (\texttt{gap\_decay\_gamma}). A
single shared setting covers both ALFWorld model scales.}
\label{tab:hyperparams}
\begin{tabular}{l c c c c c c c c}
\toprule
\textbf{Method} & $\eta$ & $G$ & $\epsilon_{\mathrm{low}}/\epsilon_{\mathrm{high}}$ & $\alpha_{\mathrm{KL}}$ & $\lambda$ & $b$ & $\gamma$ & SRS \\
\midrule
\methodname{} & $10^{-6}$ & 8 & 0.2/0.24 & 0.01 & 0.5 & 0.2 & 0.95 & KM \\
\bottomrule
\end{tabular}
\end{table*}

A single optimization recipe is shared by both ALFWorld model scales:
learning rate, group size, clipping range, and KL coefficient as given
in Table~\ref{tab:hyperparams}, plus a dual-clip constant $c{=}3.0$,
gradient clipping $1.0$, entropy coefficient $0.001$, a single PPO epoch
per update, and FSDP on one node. The remaining settings specific to
ALFWorld appear in Table~\ref{tab:env_config}.

\begin{table}[t]
\centering
\caption{\textbf{ALFWorld training configuration.} Optimization settings
already covered by the recipe in Table~\ref{tab:hyperparams} are not
repeated; only ALFWorld-specific settings are listed below.}
\label{tab:env_config}
\begin{tabular}{l c}
\toprule
 & \textbf{ALFWorld} \\
\midrule
Training steps & 150 \\
Train batch size & 16 \\
Rollout group size $G$ & 8 \\
Max prompt length & 2048 \\
Max response length & 512 \\
Max interaction turns & 50 \\
Rollout temperature (train / val) & 1.0 / 0.4 \\
GPUs (tensor-parallel size) & 8 (2) \\
\bottomrule
\end{tabular}
\end{table}

\section{Training Dynamics}
\begin{figure}[t]
\centering
\includegraphics[width=\columnwidth]{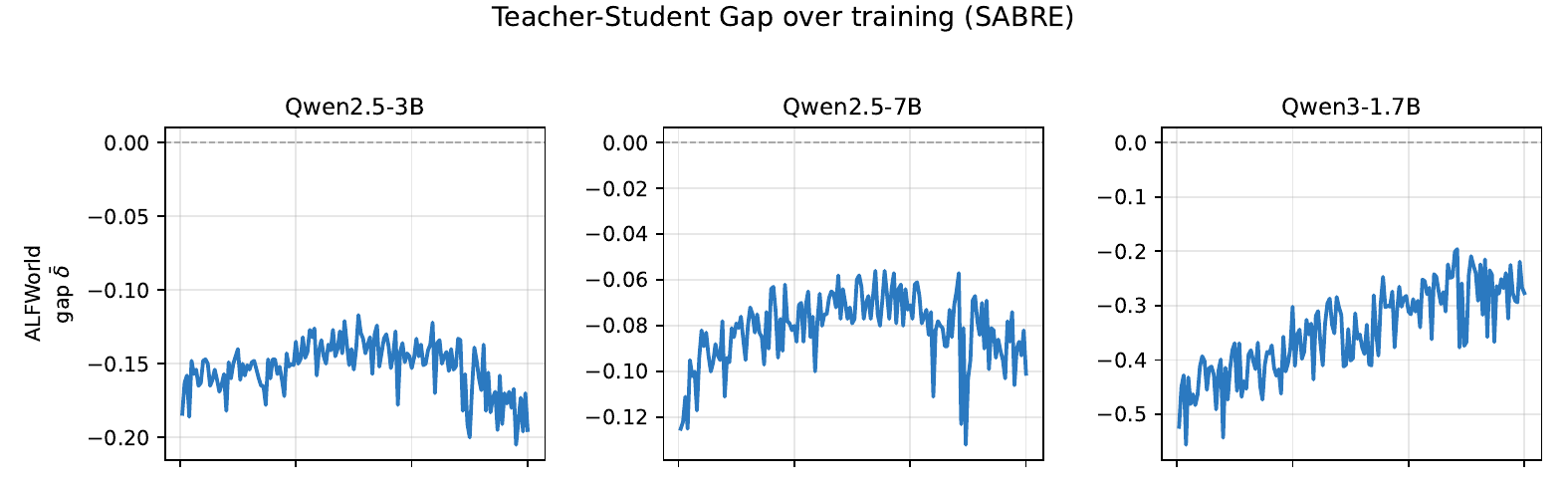}
\caption{\textbf{Teacher--Student Gap} $\bar\delta$ recorded while
training Qwen2.5-3B and Qwen2.5-7B on ALFWorld.}
\label{fig:metrics_teacher_gap_mean}
\end{figure}

\begin{figure}[t]
\centering
\includegraphics[width=\columnwidth]{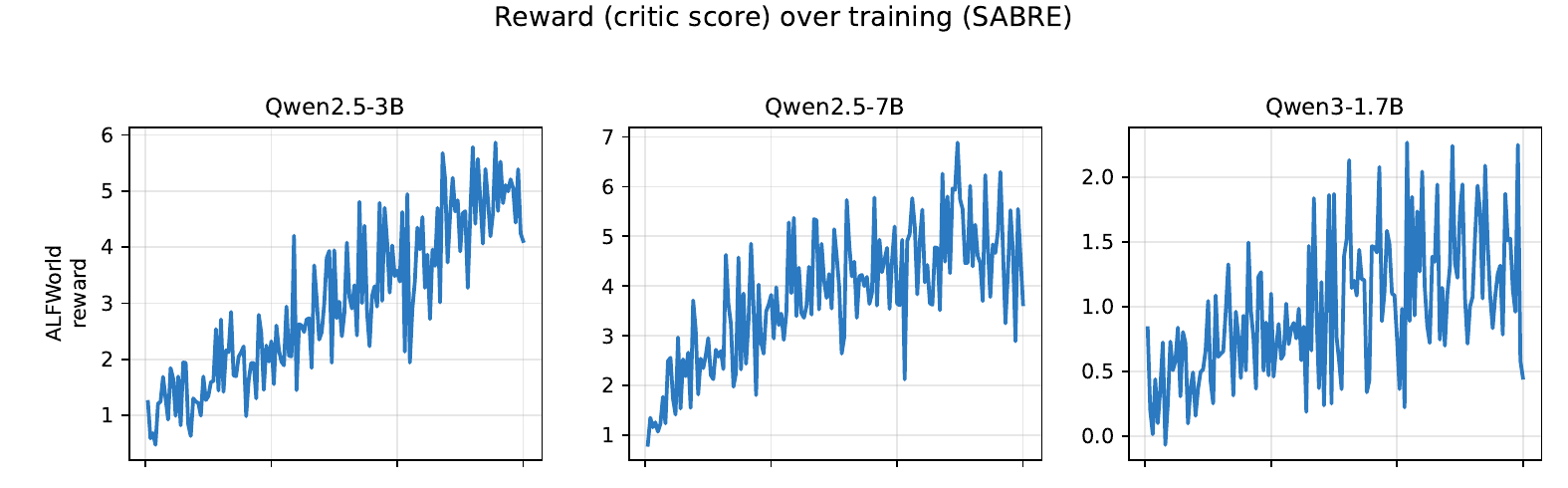}
\caption{\textbf{Reward Curve} recorded while training Qwen2.5-3B and
Qwen2.5-7B on ALFWorld.}
\label{fig:metrics_critic_score_mean}
\end{figure}

Figures~\ref{fig:metrics_teacher_gap_mean}--\ref{fig:metrics_critic_score_mean}
present the complete training dynamics for \methodname{} on ALFWorld at
both model scales, following how the teacher--student gap and the reward
evolve as training proceeds.

\clearpage

\section{Prompt}

Figure~\ref{fig:prompt_alfworld} shows the full prompt template that
\methodname{} uses on ALFWorld. The placeholder
\texttt{\{skill\_context\}} carries the retrieved skill during training
and is left empty at inference. All other placeholders are filled in
from the current interaction state, so the policy only observes its own
history together with the environment feedback.

\begin{figure*}[t]
\centering
\begin{templatebox}{Prompt of \methodname on ALFWorld}
You are an expert agent operating in the ALFRED Embodied Environment. Your task is to: \{task\_description\}.

\{skill\_context\}

Prior to this step, you have already taken \{step\_count\} step(s). Below are the most recent \{history\_length\} observations and the corresponding actions you took: \{action\_history\}

You are now at step \{current\_step\} and your current observation is: \{current\_observation\}

Your admissible actions of the current situation are: [\{admissible\_actions\}].

Now it's your turn to take an action.
You should first reason step-by-step about the current situation. This reasoning process MUST be enclosed within \texttt{<think> </think>} tags.
Once you've finished your reasoning, you should choose an admissible action for current step and present it within \texttt{<action> </action>} tags.
\end{templatebox}
\caption{The complete prompt given to \methodname{} for ALFWorld.}
\label{fig:prompt_alfworld}
\end{figure*}

%% file: tables/experiment.tex
\providecommand{\methodname}{AgentOPSD}

\definecolor{bestcellcolor}{HTML}{E2D4F0}
\definecolor{secondcellcolor}{HTML}{D9EAF7}

\newcommand{\bestval}[1]{%
    \cellcolor{bestcellcolor}\textbf{#1}%
}
\newcommand{\secondval}[1]{%
    \cellcolor{secondcellcolor}\underline{#1}%
}

\begin{table*}[t]
    \centering
    \caption{
        \textbf{Results on ALFWorld.}
        We report success rates (\%) for each of the six task categories along
        with their sample-weighted average. All methods access skills at
        training time only, apart from those tagged with $*$, which keep
        them at validation as well. \methodname{} itself requires no skills at test time.
        \sethlcolor{bestcellcolor}\hl{\textbf{Top}} and
        \sethlcolor{secondcellcolor}\hl{\mbox{\underline{runner-up}}} entries
        are marked per model scale.
    }
    \label{tab:main_results}

    \begin{tabular}{l ccccccc}
    \toprule
    & \multicolumn{7}{c}{\textbf{ALFWorld}} \\
    \cmidrule(lr){2-8}

    \textbf{Method}
    & \textbf{Pick}
    & \textbf{Look}
    & \textbf{Clean}
    & \textbf{Heat}
    & \textbf{Cool}
    & \textbf{Pick2}
    & \textbf{Avg} \\
    \midrule
    \rowcolor{gray!10}
    \multicolumn{8}{l}{\textit{Qwen2.5-3B-Instruct}} \\
    Vanilla
        & 44.4 & 11.1 & 6.2 & 15.4 & 28.6 & 12.5 & 21.9
        \\
    Skill-Prompt*
        & 51.7 & 66.7 & 48.4 & 0.0 & 4.3 & 10.0 & 28.9
        \\
    OPSD
        & 48.8 & 41.7 & 16.7 & 0.0 & 15.8 & 16.7 & 28.1
        \\
    GRPO
        & 91.2 & 62.5 & \secondval{96.2} & 61.9 & 65.0 & 47.4 & 75.0
        \\
    Skill-GRPO
        & 88.9 & \secondval{71.4} & 58.8 & \secondval{70.6} & 40.7 & 29.2 & 60.2
        \\
    Skill-GRPO*
        & \secondval{94.3} & 57.1 & \bestval{100} & 66.7 & 73.1 & 57.1 & 80.5
        \\
    GRPO+OPSD
        & \bestval{100} & \bestval{82.4} & 85.7 & \bestval{75.0} & 70.0 & 60.0 & \secondval{81.2}
        \\
    StepOPSD
        & 82.4 & 66.7 & 82.6 & 52.2 & \secondval{73.7} & \bestval{75.0} & 73.4
        \\
    \textbf{\methodname{}}
        & \bestval{100} & 68.8 & 86.4 & 66.7 & \bestval{84.6} & \secondval{73.7} & \bestval{84.4}
        \\
    \midrule
    \rowcolor{gray!10}
    \multicolumn{8}{l}{\textit{Qwen2.5-7B-Instruct}} \\
    Vanilla
        & 36.1 & 22.2 & 3.1 & 0.0 & 0.0 & 0.0 & 12.5
        \\
    Skill-Prompt*
        & 51.7 & 50.0 & 32.3 & 5.3 & 4.3 & 0.0 & 23.4
        \\
    OPSD
        & 50.0 & 60.0 & 22.7 & 21.4 & 17.6 & 9.5 & 32.8
        \\
    GRPO
        & 91.2 & \bestval{87.5} & 96.2 & 81.0 & 65.0 & 57.9 & 81.2
        \\
    Skill-GRPO
        & 88.5 & 66.7 & 65.2 & 61.1 & 57.7 & 73.1 & 69.5
        \\
    Skill-GRPO*
        & \bestval{100} & \secondval{83.3} & \secondval{96.4} & 83.3 & 75.0 & \secondval{78.9} & 88.3
        \\
    GRPO+OPSD
        & 91.4 & 61.5 & \bestval{100} & \secondval{87.5} & \secondval{76.5} & 52.2 & 80.4
        \\
    StepOPSD
        & \secondval{98.1} & 75.0 & \bestval{100} & \bestval{90.5} & \bestval{80.0} & 63.2 & \secondval{88.4}
        \\
    \textbf{\methodname{}}
        & 91.2 & \bestval{87.5} & \bestval{100} & \bestval{90.5} & 75.0 & \bestval{84.2} & \bestval{89.1}
        \\
    \bottomrule
    \end{tabular}
\end{table*}

%% file: tables/ablation.tex
\begin{table*}[t]
\centering
\begin{tabular}{l p{6.4cm} c}
\toprule
Ablated component & Alteration & ALFWorld \\
\midrule

\rowcolor{gray!10}
\textbf{\methodname{} (full)}
& bounded reshaping of turn-level credit, $\lambda{=}0.5$
& 89.1 \\

\midrule

Turn-level granularity of credit
& score tokens individually rather than turns
& 85.9 \\

Recursive revision of the belief state~\eqref{eq:belief}
& use the raw local gap $e_k$ in place of $\Delta B_k$
& 82.8 \\

Outcome-signed credit direction~\eqref{eq:credit}
& keep only the magnitude $|\Delta B_k|$ and drop the outcome sign
& 80.5 \\

Anchor at the state prior $B_0$
& initialize without the empirical rate
& 78.9 \\

\bottomrule
\end{tabular}

\caption{
\textbf{Component-wise ablation of \methodname{}}
on ALFWorld with Qwen2.5-7B; values denote success rate (\%).
Each row removes or replaces exactly one mechanism while every
other setting remains unchanged. Removing the outcome-signed
direction or the anchor on the state prior causes the largest
drop, whereas the
recursive revision of the belief state and the turn-level
granularity of credit bring smaller yet consistent improvements.
}

\label{tab:ablation}
\end{table*}

%% file: sections/proof.tex
\section{Theoretical Derivations}
\label{appendix:theory}

\subsection{Self-Teacher Contrast as a Surrogate for the Bayes Factor}
\label{app:bayes_approx}

We take the turn-wise Bayes factor as the reference quantity,
\begin{equation}
\begin{aligned}
\mathcal{B}_k
&=\log\frac{p(a_k\mid s_k,C)}{p(a_k\mid s_k,\neg C)}\\
&=\operatorname{logit}p(C\mid s_k,a_k)-\operatorname{logit}p(C\mid s_k)
\end{aligned}
\label{eq:app_bayes}
\end{equation}
while \methodname{} substitutes the self-teacher contrast in its place,
\begin{equation}
e_k \;=\; \log\frac{\pi_\theta(a_k\mid s_k,c^{+})}{\pi_\theta(a_k\mid s_k)},
\label{eq:app_ek}
\end{equation}
Throughout, $C$ stands for eventual success and $\rho_k=p(C\mid s_k)$. Two assumptions underpin the derivation:
\textbf{(A1)} the skill-conditioned branch behaves as the success-conditional policy,
$\pi_\theta(a_k\mid s_k,c^{+})\approx p(a_k\mid s_k,C)$;
\textbf{(A2)} in the rare-success regime ($\rho_k$ small), failed rollouts dominate the marginal,
$\pi_\theta(a_k\mid s_k)\approx p(a_k\mid s_k,\neg C)$.

The marginal action distribution mixes the success- and failure-conditioned policies,
\begin{equation}
\pi_\theta(a_k\mid s_k)=\rho_k\,p(a_k\mid s_k,C)+(1-\rho_k)\,p(a_k\mid s_k,\neg C).
\label{eq:app_mixture}
\end{equation}
Substituting \eqref{eq:app_mixture} into \eqref{eq:app_ek} under (A1) gives
\begin{equation}
\begin{aligned}
e_k &\;\approx\; \mathcal{B}_k-\log\!\big(1-\rho_k+\rho_k\,e^{\mathcal{B}_k}\big)\\
&\;\xrightarrow[\;\rho_k\to0\;]{}\; \mathcal{B}_k ,
\end{aligned}
\label{eq:app_bias}
\end{equation}
Hence (A2) isolates the $\rho_k\to0$ regime, where the contrast coincides with the Bayes factor. Under (A1) alone, on the other hand, $e_k$
reduces to the pointwise mutual information
\begin{equation}
e_k \;\approx\; \log\frac{p(a_k\mid s_k,C)}{p(a_k\mid s_k)}
\;=\; \log\frac{p(C\mid s_k,a_k)}{p(C\mid s_k)} ,
\label{eq:app_pmi}
\end{equation}
which is positive if and only if $a_k$ increases the posterior probability of success. In \eqref{eq:app_bias}, the correction is
monotone in $\mathcal{B}_k$; it follows that $\operatorname{sign}(e_k)=\operatorname{sign}(\mathcal{B}_k)$, meaning $e_k$
preserves the ordering of turns by evidential strength. \methodname{} consults $e_k$ solely through this sign and ordering.

\subsection{Properties of the Reshaped Advantage}
\label{app:reshape_props}

Let $A^{(i)}$ denote the group-relative advantage, and let $\Delta B_k=B_k-B_{k-1}$ denote the per-turn belief revision. We further write $z_k$
for its within-trajectory standardization,
$m_k=\mathrm{clip}\!\big(1+b\,\operatorname{sign}(A^{(i)})z_k,\,1-b,\,1+b\big)$ for the bounded multiplier with $b\in(0,1)$, and
$\tilde A_k=A^{(i)}\big((1-\lambda)+\lambda m_k\big)$ for the reshaped advantage with $\lambda\in[0,1]$.

\begin{proposition}[Bounded Perturbation]\label{prop:bounded}
$\big|\tilde A_k-A^{(i)}\big|\le\lambda b\,|A^{(i)}|$, so that
$(1-\lambda b)|A^{(i)}|\le|\tilde A_k|\le(1+\lambda b)|A^{(i)}|$.
\end{proposition}
\begin{proof}
$m_k\in[1-b,1+b]$ implies $|m_k-1|\le b$, while $\tilde A_k-A^{(i)}=A^{(i)}\lambda(m_k-1)$.
\end{proof}

\begin{proposition}[Preservation of Sign]\label{prop:sign}
$\operatorname{sign}(\tilde A_k)=\operatorname{sign}(A^{(i)})$ holds at each turn $k$.
\end{proposition}
\begin{proof}
$(1-\lambda)+\lambda m_k\ge 1-\lambda b>0$ because $\lambda\le1,\,b<1$; multiplication by a strictly positive factor leaves the sign unchanged.
\end{proof}

\begin{proposition}[GRPO as a Special Case]\label{prop:recover}
When $\lambda=0$, $\tilde A_k=A^{(i)}$ at every token, and the \methodname{} gradient reduces to the GRPO gradient.
\end{proposition}
\begin{proof}
Setting $\lambda=0$ yields $(1-\lambda)+\lambda m_k=1$; hence $\tilde A_k=A^{(i)}$ identically, whatever the belief signal.
\end{proof}

\begin{proposition}[First-Order Expansion of the Belief Revision]\label{prop:decomp}
With $c_k=\gamma c_{k-1}+e_k$, $\ell_k=\operatorname{logit}(B_0)+c_k$, and $\Delta\ell_k=e_k-(1-\gamma)c_{k-1}$, we have
\begin{equation}
\Delta B_k=B_{k-1}(1-B_{k-1})\,\Delta\ell_k+O\!\big((\Delta\ell_k)^2\big).
\end{equation}
\end{proposition}
\begin{proof}
$B_k=\sigma(\ell_k)$, where $\sigma'=\sigma(1-\sigma)$; Taylor expansion to first order about $\ell_{k-1}$ yields
$B_k=B_{k-1}+B_{k-1}(1-B_{k-1})\Delta\ell_k+O((\Delta\ell_k)^2)$. The gating factor $B(1-B)$ peaks at $B=\tfrac12$
and tends to zero as $B\to\{0,1\}$.
\end{proof}

\begin{proposition}[Telescoping Property of the Idealized Recursion]\label{prop:budget}
Summing the idealized increments telescopes to the net change, $\sum_k\Delta B_k=B_K-B_0$.
\end{proposition}
\begin{proof}
$\sum_{k=1}^{K}(B_k-B_{k-1})=B_K-B_0$.
\end{proof}

\begin{proposition}[Non-Identifiability of Turn-Level Contributions]\label{prop:unidentifiable}
One can find two trajectories sharing the same outcome reward---and therefore the same broadcast advantage---whose
per-turn contributions nevertheless differ; turn-level credit is thus not recoverable from the trajectory's return by itself.
\end{proposition}
\begin{proof}[Proof (by construction)]
Consider $\tau_1,\tau_2$ within a single group satisfying $R(\tau_1)=R(\tau_2)$, whence $A(\tau_1)=A(\tau_2)$ and GRPO broadcasts one
scalar to all turns. Suppose $\tau_1$ succeeds via one decisive turn ($\Delta B$ concentrated), whereas
$\tau_2$ succeeds via progress spread evenly across turns. Although the returns agree, the underlying per-turn contributions do not,
so recovering them calls for an extra per-turn signal.
\end{proof}

\begin{proposition}[$B_0$ as an Estimate of the Group Success Rate]\label{prop:v0}
Consider a task whose success probability is $\theta_x$, together with a group of $G$ trajectories, of which $S$ succeed given
a binary reward. The maximum-likelihood estimator of $\theta_x$ equals the group success fraction $\bar R=S/G$,
i.e., the usual GRPO group mean; accordingly, \methodname{} initializes $B_0=\operatorname{clip}(\bar R,\epsilon_0,1-\epsilon_0)$.
\end{proposition}
\begin{proof}
The MLE under a Binomial$(G,\theta_x)$ likelihood equals $S/G$. Clipping ($\epsilon_0{=}10^{-4}$) merely keeps
$\operatorname{logit}(B_0)$ finite for groups that are entirely correct or entirely wrong; such groups carry zero group-relative
advantage and therefore contribute nothing to the update.
\end{proof}

%% file: sections/algorithm.tex
\section{Training Algorithm}
\label{appendix:algorithm}
Algorithm~\ref{alg:sabre} summarizes one \methodname{} training iteration at the granularity of turns. Relative to GRPO, the added components are limited to one teacher forward pass at every turn, along with the belief-reshaping block applied per turn; all remaining steps follow the standard group-relative update.

\begin{algorithm*}[t]
\caption{\methodname{}: Turn-Level Credit Assignment via Recursive Belief Updates}
\label{alg:sabre}
\begin{algorithmic}[1]
\Require policy $\pi_\theta$, verifier of rewards $R$, group size $G$, plus a skill retriever; mixing coefficient $\lambda$, clipping range $b$, evidence discount $\gamma$
\For{every training iteration}
  \State sample a mini-batch of tasks $\{x\}$
  \For{each task $x$ together with its retrieved skill $c^{+}$}
    \State draw $G$ trajectories $\{y^{(1)},\dots,y^{(G)}\}\sim\pi_\theta(\cdot\mid x)$; each trajectory $i$ contains $K_i$ turns \Comment{sample on-policy rollouts}
    \For{$i=1,\dots,G$}
      \State obtain from the verifier the reward $R^{(i)}=R(x,y^{(i)})\in\{0,1\}$
    \EndFor
    \State $A_{\mathrm{seq}}^{(i)}\gets (R^{(i)}-\mu_G)/\sigma_G$ \Comment{group-relative advantage estimate}
    \For{$i=1,\dots,G$}
      \State $B_0\gets\mathrm{clip}(\bar R,\epsilon_B,1-\epsilon_B)$;\quad $\ell_0\gets\mathrm{logit}(B_0)$ \Comment{initialize the prior using the empirical success fraction $\bar R{=}S/G$ (Prop.~\ref{prop:v0})}
      \State $c_0\gets 0$
      \For{$k=1,\dots,K_i$} \Comment{iterate over turns, evolving the belief state}
        \State $e_k\gets\sum_t\mathrm{sg}[\log\pi_\theta(y_{k,t}\mid s_k^{+})-\log\pi_\theta(y_{k,t}\mid s_k)]$ \Comment{one extra teacher forward pass}
        \State $c_k\gets\gamma\,c_{k-1}+e_k$;\quad $\ell_k\gets\ell_0+c_k$;\quad $B_k\gets\sigma(\ell_k)$;\quad $\Delta B_k\gets B_k-B_{k-1}$
      \EndFor
      \State $q_k\gets\mathrm{sign}(A_{\mathrm{seq}}^{(i)})\,\Delta B_k$ for all $k$ \Comment{sign credit by the trajectory outcome}
      \State $z_k\gets (q_k-\mathrm{mean}(q))/(\mathrm{std}(q)+\epsilon)$ \Comment{standardize per trajectory}
      \State $w_k\gets\mathrm{clip}(1+b\,z_k,\,1-b,\,1+b)$ \Comment{clip into a bounded multiplier}
      \State $\widetilde A_k^{(i)}\gets A_{\mathrm{seq}}^{(i)}\big[(1-\lambda)+\lambda w_k\big]$; each token takes the $\widetilde A$ of the turn it belongs to
    \EndFor
  \EndFor
  \State optimize $\theta$ on the clipped GRPO objective $\mathcal{L}_{\methodname}(\theta)$ using $\{\widetilde A\}$ \Comment{perform the policy-gradient update}
\EndFor
\end{algorithmic}
\end{algorithm*}

\paragraph{Token-level variant.} In the token-level variant, the same recursion runs over the response-masked, flattened stream of tokens rather than over turns: $\delta_t$ accumulates directly at token level, $\Delta B_t$ undergoes standardization across all tokens of a trajectory, and $\widetilde A_t$ is assigned token by token. This variant appears in Table~\ref{tab:ablation} as the granularity ablation.

\paragraph{Cost.} Compared with GRPO, the only added overhead is one teacher forward pass per trajectory. Since the belief-reshaping block acts elementwise, it introduces neither extra rollouts nor any trainable parameters.